# Robust Ship Detection and Tracking Using Modified ViBe and Backwash Cancellation Algorithm

Mohammad Hassan Saghafi[1] , Seyed Abolfazl Seyed Javadein[2], Seyed Majid Noorhosseini[3], Hadi Khalili[4]

[1] Department of Electrical Engineering,
Amirkabir University of Technology, Tehran, Iran
[2] Member of National Elites Foundation of Iran
[3] Department of Computer Engineering,
Amirkabir University of Technology, Tehran, Iran
[4] Bina Intelligent Machatronics Inc, Tehran, Iran
`{Mohamadsaghafi, Majidnh}@aut.ac.ir` ,
`Seyedjavadein@negareh.ac.ir` , `Khalilih@binaco.ir`

**Abstract.** In this paper, we propose a robust real time detection and tracking method for detecting ships in a coastal video sequences. Since coastal scenarios are unpredictable and scenes have dynamic properties it is essential to apply detection methods that are robust to these conditions. This paper presents modified ViBe for moving object detection which detects ships and backwash. In the modified ViBe the probability of losing ships is decreased in comparison with the original ViBe. It is robust to natural sea waves and variation of lights and is capable of quickly updating the background. Based on geometrical properties of ship and some concepts such as brightness distortion, a new method for backwash cancellation is proposed. Experimental results demonstrate that the proposed strategy and methods have outstanding performance in ship detection and tracking. These results also illustrate real-time and precise performance of the proposed strategy.

## 1 Introduction

Detecting moving objects in a video sequence is a fundamental and critical task in automatic video content analysis especially in video surveillance and tracking. Numerous methods for Moving objects detection have been proposed over the years. Among different objects detection methods that have been considered in various works, ships detection and coastal scenarios have attracted less attention. However, the important applications of ship detection in coastal surveillance, ship tracking and counting the number of passing ships and boats have stimulated scientists to pay more attention to this topic recently. Automated ship detection is a challenging problem, because of the unpredictable environment and high requisites for robustness and accuracy.

In this paper we consider that we have mounted a stationary camera and there are different ships and vessels on coastal environment. We present an automated ship detection system that is robust to backwash in behind of ships and natural waves that can be seen in non-stormy sea. In the proposed strategy we detect the ships in every frame and then extract important information about the region that includes the ship. By using this precise information, we can track the ships accurately.

The most primary task in detection is to determine moving objects like ships and waves which are caused by ship's movement (backwash). In cases where the camera is fixed, various methods for moving objects detection can be divided into three groups.

The first group is known as 'temporal difference'. When the camera is fixed the method has several major problems like fragmentation and slow moving object detection. Optical flow is used in the second group. This method is not appropriate to scan the entire frame and real time detecting all moving objects. The third group of foreground detection is categorized as background subtraction. According to [2] which surveyed different methods of the third group, we can conclude that ViBe [3] has more advantages over other methods.

We studied the ViBe algorithm, and used it with some modifications to implement the first step of ship detection strategy in section 2 of this paper. After detecting moving objects, ship and backwash, it is essential to make the moving objects more clear. Toward this end, removing noises, and then morphological processing are done in section 3. Determination of connected components is also illustrated in section 3. In remainder of this paper, our method about backwash cancellation and determining of regions included ships is expressed in section 4. Experimental results are presented in section 5. Section 6 concludes the paper. Fig.1 shows the flowchart and architecture of the proposed method.

## 2   Background Subtraction: Modified ViBe Algorithm

ViBe algorithm [3] was proposed by O. Barnich and M.V. Droogenbroeck, builds background models which is used to get foreground at every moment. ViBe, a powerful random strategy, could cope with multi modes pixel background same as GMM [3] and other advanced techniques method but faster and simpler than them, so ViBe can identify small natural waves as background in sea scenario. In addition if a serious change occurs in a part of background that leads to new scene such as adding or eliminating a fixed object to background, ViBe can handle it.

### 2.1   Pixel Model

The ViBe algorithm is based on pixel model. In contrary parametric methods such as [1] that consider a probability density function as pixel model, ViBe defines pixel model as a set of values that each pixel might possesses as a background pixel in different frames. Whenever a model is built for a pixel by using previous frames, thereafter, each pixel related to that pixel can be classified as foreground or background at current frame. To classify p(x,t), a pixel value at time t and position x, it is compared with members of related pixel model to it. p(x,t) is identified as background, if there are at least $\phi_{min}$ members that difference between them and p(x,t) be lower than a predefined threshold (R), otherwise it is classified as foreground. Our contribution in ViBe is about considering two predefined threshold. Each frame is divided to two regions. We classify pixels that have more probability to become a part of foreground, as first region and other pixels of frame is classified as second region. We define a small threshold (R1) for first region and a higher threshold (R2) for second region. By this innovation, it is possible to detect objects when the background has a similar color to objects in the scene. Determining of first region is done at the last stage of our strategy; it is observable in Fig. 1.

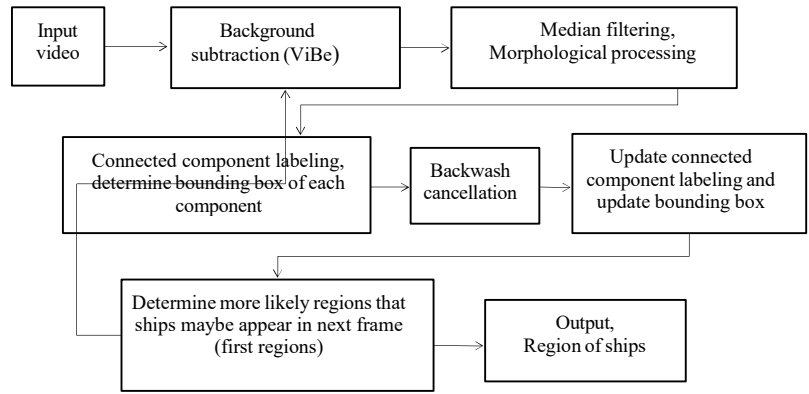

**Fig. 1.** Flowchart of proposed method

### 2.2 Initializing and Updating the Model

There are various strategies to initialize the model; if it is done correctly we can get an accurate background estimate as soon as possible. Since we have assumed 20 members for each model, we applied 20 first frames to initial pixels model. Although in [3] proposes to use first frame for initializing the models but we get superior responses by using 20 first frames, because different permanent modes of each pixel are stored in pixel model.

The background is changing over time because of variations in illumination and background geometry, so updating the models is inevitable over time. There is a precondition to find whether the model of a pixel needs to update. The precondition is established for a pixel if it is classified as background pixel at current frame. A simple rule for updating can be illustrated as: if precondition is established for a pixel, then related model to this pixel must be updated. But ViBe has better performance, if updating is performed with a chance whenever precondition is established. In order to adding the value of background pixel to related model to it, selecting a member of model for removing is another important point in updating. Experiments have shown that selecting randomly a member of pixel model to eliminate gives better responses than any other selecting strategies. In ViBe when each pixel model is updated, a neighboring pixel also is randomly chosen to update its pixel model. With this trick we can solve some problematic cases like when a serious change in portion of background occurs. For example when a ship drops anchor in harbor to stop for time duration. In addition, by using the neighboring pixel updating, ViBe can admit multi modes pixel as background pixel if different modes of that pixel exist in neighboring pixels. For instance when background consists of small naturally waves in sea scenario.

## 3   Removing Noises and Connected Component Labeling

There always exist some small noise areas both in the foreground and background regions, so we use a 3*3 size of median filter to remove and minimize the noises and worthless data. After determination of moving objects and removing noise, ship and its backwash seem fragmented. So we use morphological operations to fill discontinuities of each moving object. In this paper, the close operation of morphological processing is applied two times for each frame.

Once moving objects is obtained, we utilize RLC [4] that is a new component-labeling method for clustering different moving objects in each frame. RLC belongs to fast connected component labeling group so it is good for real time applications such as ship tracking. After clustering, minimum bounding box is determined around each cluster that contains ship and backwash, this bounding box is used as input to backwash cancellation algorithm.

## 4   Backwash Cancellation and Determining of Regions Included Ships

We propose an efficient method to remove backwash based on difference between height of ship and backwash in their bounding box and also distinction in gray scale of ship and backwash. In bounding box, some parts are smaller than half-height of bounding box so they belong to backwash and they should be removed of bounding box. In the cases that ship and backwash are both in bounding box, removing by height is suitable for backwash cancellation. But in the cases that just backwash exists on bounding box or there are two ships that are connected by backwash in bounding box, removing backwash by height is not useful. In these cases it is tried to remove backwash by using some differences between ship and its backwash such as color, texture, and brightness and so on. Brightness distortion is defined as difference between projection of RGB vector of a pixel in current frame and RGB vector of related pixel in background. The brightness distortion values for backwash are well defined in range [*T1, T2*] so it is possible to eliminate backwash. The results are improved by evaluating photometric gain (ratio between $B_n(x, y)$ and $I_n(x, y)$). According to the histogram of gain matrix, a domain [*Tp1, Tp2*] is determined to detect remainder of backwash. So we can eliminate backwash by using of brightness distortion and photometric gain. After elimination of backwash, updating connected component labeling is done and then new bounding boxes are determined that contain only ships.

## 5   Experimental Results

The performance of proposed method was evaluated using experiments. The algorithms were implemented in C++ using the OpenCv libraries in order to achieve real-time tracking performance. Experiments were conducted on a computer with an Intel Core(TM) i7, 2GHz and 6GB of ram.

Our method was tested on videos that were provided by [5], [6]. Our experiments have shown, the average execution time was only about 3ms for each frame which has a size of 200*150 RGB pixels and about 10ms for each frame which has a size of 640*480 RGB pixels. Fig.2 shows the results of our method in ship detection and backwash cancellation .We defined a validation function (*VF*) to assess accuracy of our algorithm in ship detection and backwash cancellation. This function is illustrated in (1).

$$VF = 1 - \frac{0.5\,ND + SND}{SP}. \qquad (1)$$

In (1), *SP* is defined as number of frames that ships exist on them. *ND* is number of frames that in addition ships, noises are detected as moving object. *SND* also is number of frames that ships exist on them but they were not detected by detection algorithms. Table 1 depicts the skill of the proposed algorithm to detect ships truly by using of *VF* for different video sequences. (In this paper, occlusions between ships are not considered. So the frames that included occlusions are neglected in experimenting the input video titled 'i').

## 6   Conclusions

A new method for backwash cancellation was proposed in this work. Furthermore a powerful background subtraction method based on ViBe was proposed to detect moving objects. The capacity of the proposed algorithms in real time and robust ship detection and tracking was validated by experimental study. In the future work, we will try to propose a combination of our strategy and a predictor to improve the tracking performance of the proposed strategy in the cases that there is an occlusion between two ships.

**Table 1.** Validation of proposed method

| Input Video | Size    | SP   | ND | SND | VF % | Execution time per frame(ms) |
|-------------|---------|------|----|-----|------|------------------------------|
| a           | 200*150 | 1150 | 8  | 0   | 99%  | 3                            |
| e           | 640*480 | 1900 | 6  | 3   | 99%  | 10                           |
| i           | 200*150 | 1050 | 70 | 0   | 96%  | 3.6                          |
| m           | 200*150 | 200  | 0  | 0   | 100% | 2.8                          |

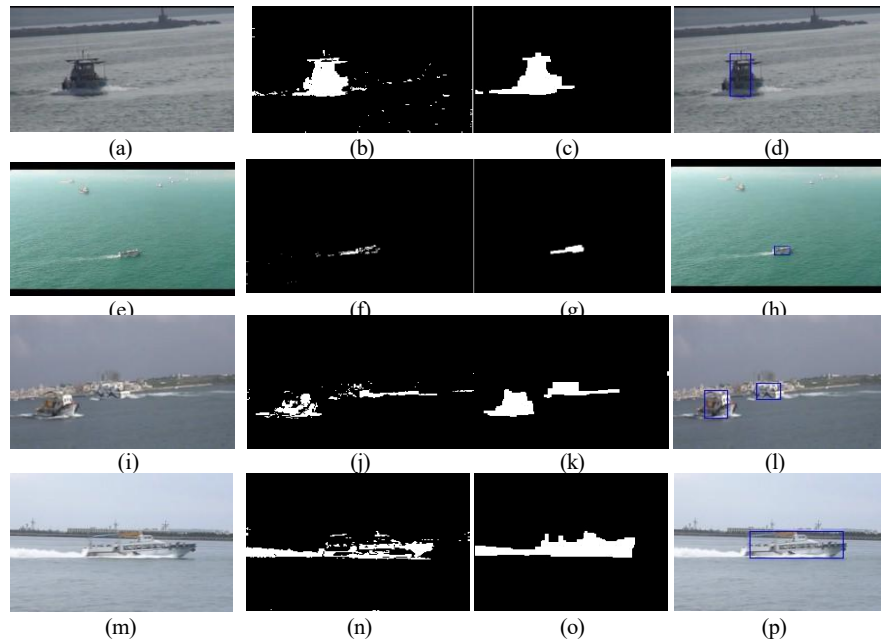

**Fig. 2.** Detecting ships: (a ,e ,i ,m) Original fames; (b ,f ,j ,n) Results of modified ViBe; (c ,g ,k ,o) frames after connected component labeling stage; (d ,h ,l ,p) Final results after backwash cancellation.